\documentclass[preprint,review,times,10pt]{elsarticle}

\usepackage{amssymb}
\usepackage{amsmath}
\usepackage{makecell} 
\usepackage{multirow}  
\usepackage[T1]{fontenc}
\usepackage{xcolor}

\begin{document}
\biboptions{numbers,sort&compress}
\begin{frontmatter}



\title{Parse Graph-Based Visual-Language Interaction for Human Pose Estimation}


\author{Shibang Liu} 
\author{Xuemei Xie} 

\author{Guangming Shi} 


\begin{abstract}
Parse graphs boost human pose estimation (HPE) by integrating context and hierarchies, yet prior work mostly focuses on single modality modeling, ignoring the potential of multimodal fusion. Notably, language offers rich HPE priors like spatial relations for occluded scenes, but existing visual-language fusion via global feature integration weakens occluded region responses and causes alignment and location failures. To address this issue, we propose Parse Graph-based Visual-Language interaction (PGVL) with a core novel Guided Module (GM). In PGVL, low-level nodes focus on local features, maximizing the maintenance of responses in occluded areas and high-level nodes integrate global features to infer occluded or invisible parts. GM enables high semantic nodes to guide the feature update of low semantic nodes that have undergone cross attention. It ensuring effective fusion of diverse information. PGVL includes top-down decomposition and bottom-up composition. In the first stage, modality specific parse graphs are constructed. Next stage. recursive bidirectional cross-attention is used, purified by GM. We also design network based on
PGVL. The PGVL and our network is validated on major pose estimation datasets. We will release the code soon.


\end{abstract}




\end{frontmatter}



\section{Introduction}
\label{Introduction}
Human pose estimation (HPE)~\cite{AnatPose,LRHPE,Adversarial,StructureAware} is the task of locating the precise locations of human joints in images to interpret body poses. It is widely used for other tasks (e.g., action recognition~\cite{LGAR}, asecurity monitoring and augmented reality/virtual reality). 

\begin{figure}[t]
	\centering
	\includegraphics[width=0.98\linewidth]{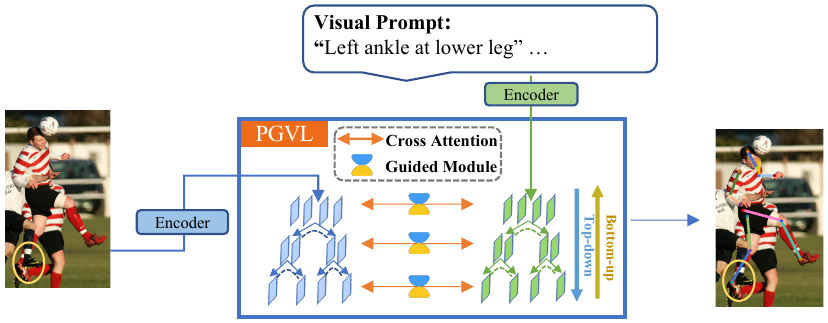}
	\caption{PGVL includes in two stages. The top-down decomposition stage converts visual and language tokens along channels into unimodal parse graphs (visual and language), respectively. The bottom-up optimization stage filters context information (horizontal dashed lines) within parse graphs and cross-modal attention via the GM module, then performs recursive computation from leaf to root nodes to generate final interactions results.}
	\label{fig1}
\end{figure}

When observing a person, humans naturally decompose the body into a hierarchical structure, from the whole to parts and then to primitives, to finish HPE better. This decomposition can be effectively represented using parse graphs~\cite{GrammarOfImages,RMPG,PGBS}, which capture hierarchical relations and context relations between body parts~\cite{PGBS} or sub-feature maps~\cite{RMPG}, enabling robust reasoning under challenging conditions such as occlusions. Additionally, humans can process and interpret visual information more quickly and accurately when provided with textual cues~\cite{LanguageNotJustTalking,SeeingObjectsThroughLanguageGlass,WordsVision}. Similarly, language prompts enhance pose estimation performance by providing spatial localization information and high-level semantic representations to better localize and distinguish different joints, especially occlusion. Recent studies have demonstrated the effectiveness of visual-language fusion in enhancing pose estimation tasks. For example, LAMP~\cite{LAMP} leverages textual supervision to improve human pose estimation accuracy, particularly in handling occlusions, by establishing cross-modal alignment from instance-level to joint-level features. Similarly, CLAMP~\cite{CLAMP} introduces the joint's prompt to animal pose estimation, further improving the performance of pose estimation. However, they use global features in multimodal fusion, which is the entire feature map. Such fusion methods reduce the response of the occluded area and cause the alignment and location to fail. In this context, the method proposed in RMPG~\cite{RMPG} offers a valuable insight. It designs the parse graph of feature map which decomposes the feature map into a parse graph with a hierarchical structure, where high level nodes contain global information and low level nodes focus on local details. Unfortunately, this approach has not been applied to multimodal fusion, where language support can significantly boost model performance, particularly in occlusion scenes.

To address the problem of alignment and location failure in occluded areas, we propose a novel method of Parse Graph-Based Visual-Language interaction (PGVL) for HPE (see Fig.~\ref{fig1}). A novel Guided Module (GM) is designed to embed into the parse graph of feature map, using high-level nodes to guide the feature update of low-level nodes that have undergone cross attention, ensuring more accurate information flow in the parse graph. PGVL includes two stage. In the {\bf{top-down decomposition stage}}, the language and visual features are separately decomposed along the channel to construct their respective parse graphs, thereby obtaining hierarchical feature structures, where the low-level layers focus on local features to prevent occlusion responses from being weakened globally, and the high-level layers focus on global features to help inference and prediction. In the {\bf{bottom-up combination stage}}, cross-attention operations~\cite{Vilbert} are first performed on the corresponding leaf node features of the two parse graphs and retain the context relations within parse graphs, after that the result are filtered by GM using the higher-level nodes. This process is iteratively repeated until reaching the root node, forming new representations for each modality. The above constitutes the complete process of PGVL. Furthermore, variants of PGVL in different semantic spaces are explored and used to build our network. While our method is primarily designed for HPE, we also evaluate its applicability to animal pose estimation. In summary, the contributions of this paper are as follows:
\begin{itemize}
	\item We propose PGVL and its variants, a novel parse graph-based visual-language interaction method that combines local and global feature advantages, and addresses the interaction and alignment problem under occlusion.
	\item A Guided Module (GM) is designed within the PGVL to effectively guide the update of node features of the parse graph and improve model performance.
	\item We develop a pose estimation network based on PGVL and validate both the network and PGVL on mainstream pose estimation datasets.
\end{itemize}

\section{Related work}
\label{Related work}

{\bf{Parse graphs.}} The parse graph includes a tree-structured decomposition and context relations among nodes~\cite{GrammarOfImages,PGBS,RMPG}. 
Parse graphs are often used to parse objects or scenes to better complete various visual tasks. This method can enhance the model's understanding of the structural features of objects or scenes, which has attracted the attention of numerous related studies. For instance, works like~\cite{PGBS,DeepFullyConnected,DeeplyLearnedCompositionalModels,HPETreeStructure,SkeletalHeatmaps} use the human body hierarchy to finish HPE better. However, its application in object parsing is prone to limitations. The method depends on a rigid framework, specifically the explicit reasoning of the parse graph of targets, and are strict data requirements. For example, PGBS~\cite{PGBS} and DFCP~\cite{DeepFullyConnected} use additional body and part labels for supervision, and SH~\cite{SkeletalHeatmaps} and DLCM~\cite{DeeplyLearnedCompositionalModels} uses additional part labels for supervision. Due to these constraints, it is difficult to integrate this method flexibly with others, thus restricting its scalability. Recent research RMPG~\cite{RMPG} proposes treating feature maps as objects (e.g., human bodies) for parsing to obtain the parse graph of feature map, thereby optimizing feature maps. The parse graph is an implicit structure that requires no additional supervision and can address the above issues to some extent. We apply parse graph of feature map to multimodal fusion and design a novel GM within the parse graph to guide node feature updates.

{\bf{ Vision-language models.}} Language is proven effective in learning visual representations for downstream tasks ~\cite{VLN,VQA,PIM,Mirrorgan}. Vision-language pretraining (typically involving joint training of image and text encoders)  acquires significant progress ~\cite{Cloob,VLNTS,CLIP}. A typical example is CLIP (Contrastive Language-Image Pretraining) ~\cite{CLIP}, which uses 400 million image-text pairs to pretrain a multimodal model with strong visual-language representation and transfer learning capabilities. How to effectively adapt these pretrained models to downstream tasks has been widely explored by researchers. In the field of human action recognition, PeVL ~\cite{PeVL} proposes a multi-level contrastive learning framework for vision-pose-language alignment. In pose estimation, LAMP ~\cite{LAMP} improves pose estimation results (especially in occluded scenarios) by introducing joint and instance descriptions, while CLAMP ~\cite{CLAMP} proposes a cue-based contrastive learning scheme to enhance animal pose estimation. These methods~\cite{CLAMP,LAMP,PeVL}, which leverage CLIP’s powerful visual-language encoding capabilities to achieve good performance on tasks, merely adopt the mainstream approach of directly using global feature map information in vision-language fusion without deeply exploring the fusion mechanism. While we split the feature map and apply cross attention to each split part so that the features of the occluded area will not be weakened by the global features.
\section{Method}
\label{Method}
In this section, we first introduce the parse graph, followed by the overall network architecture. We then describe the specific design of PGVL and the Visual-Language Matching Loss (VLML).

\subsection{Parse Graph}
\label{Parse Graph}
The parse graph includes hierarchical structures and context relations, represented as a 4-tuple $(\mathcal{V},\mathcal{E},\Phi^{and},\Phi^{leaf})$~\cite{GrammarOfImages}. Here $(\mathcal{V},\mathcal{E})$ defines the hierarchical structure, corresponding to the first stage in PGVL, while $(\Phi^{and},\Phi^{leaf})$ are potential functions. Each node $v\in \mathcal{V}$ has state variable $s_v=\{p_v,t_v\}$, where $p_v$ is the position and $t_v$ is the type. Given an image $I$, the probability of the state variables $\Theta$ is:
\begin{equation}
	P(\Theta|I) = \frac{1}{Z}\exp\{ -\Psi(\Theta, I)\}
\end{equation}
where $\Psi(\Theta, I)$ is the energy function, $Z$ is the partition function, and the energy function $F(\Theta)=-\Psi(\Theta, I)$ is decomposed as:
\begin{equation}
	F(\Theta) = \sum_{v\in \mathcal{V}_L} \Phi^{leaf}_v(s_v, I) + \sum_{u \in \mathcal{V}_A} \Phi^{and}_u(s_u, \{s_x\}_{x\in ch(u)})
\end{equation}
where $\mathcal{V}_L$ and $\mathcal{V}_A$ are leaf and non-leaf nodes respectively, and $ch(u)$ denotes the children of node $u$. The usually computation of optimal state $\Theta^*$  is carried out in two stages: bottom-up activation and top-down refinement~\cite{DeeplyLearnedCompositionalModels,PGBS,GrammarOfImages}. The bottom-up stage computes the maximum score $F_u^{\uparrow}(s_u)$, while the top-down stage refines each node $v$ using its parent node $u$ and siblings:
\begin{equation}
	\label{eq0}
	F_v^{\downarrow}(s_v) = \Phi_{u,v}(s_u^*, s_v) + \mathcal{N}_v(s_v, \{s_h\}_{h \in S_v}) 
\end{equation}
where $S_v$ contains all nodes at the same level as $v$, $\mathcal{N}_v$ captures the context relation of node $v$, and $s_u^*=\arg\max_{s_u}F_u^{\uparrow}(s_u)$. This two-stage process ensures accurate object's part predictions by leveraging hierarchical and context information. Our proposed PGVL method, based on the parse graph, introduces the cross-modal fusion mechanism and the GM. Since it operates on feature maps rather than objects, each part is known, and it only requires splitting the feature maps and optimizing them in a bottom-up composition.

\begin{figure}[t]
	\centering
	\includegraphics[width=0.98\linewidth]{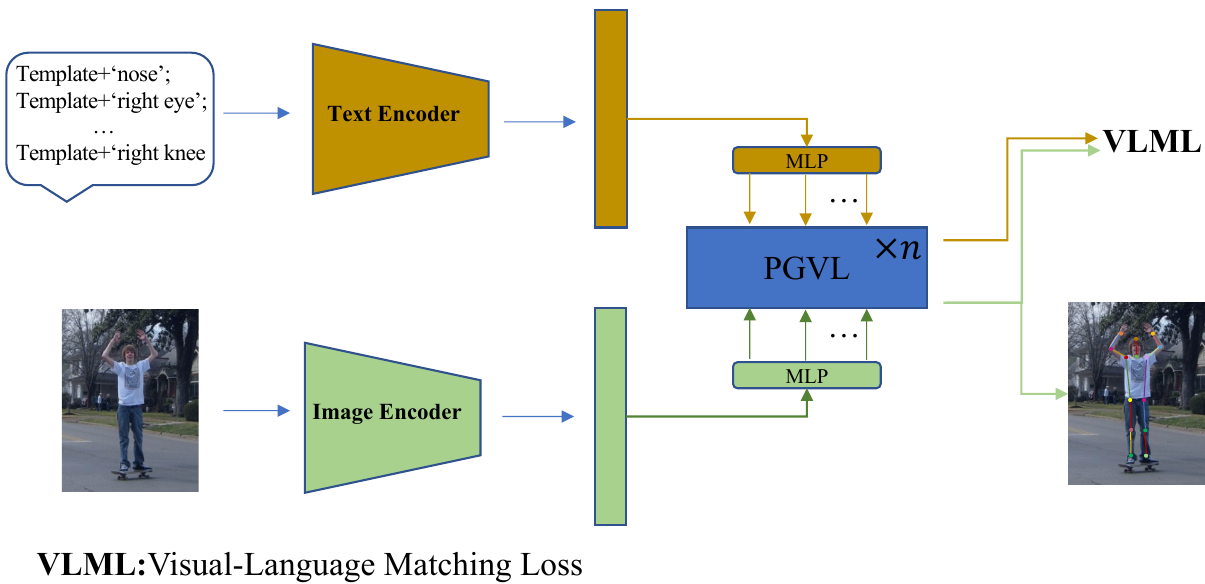}
	\caption{Our network. Images and text are encoded through their respective encoders to get their own tokens. Tokens are processed by PGVL or its variants to obtain the result of multimodal fusion. The processed visual and language tokens is used for Visual-Language Matching Loss (VLML) to make each joint visual feature closer to its language description feature. In addition, the processed visual token are also used to predict the position of the joints in the image.}
	\label{fig2}
\end{figure}

\subsection{Our Network}
{\bf{Framework.}} As shown in Fig.~\ref{fig2}, our network framework includes three parts. The first part is the encoding of vision and language. The second and most important part is visual-language fusion through the PGVL module or it's variants. The third part is VLML and the joint's mean squared error loss.

{\bf{Language embedding.}} The language embedding leverages learnable prefix templates~\cite{CLAMP} to adapt CLIP for pose estimation. Specifically, it generates grounded text prompts (e.g., 'left wrist') by populating predefined templates with joint names.

\subsection{PGVL}

\begin{figure}[t]
	\centering
	\includegraphics[width=0.9\linewidth]{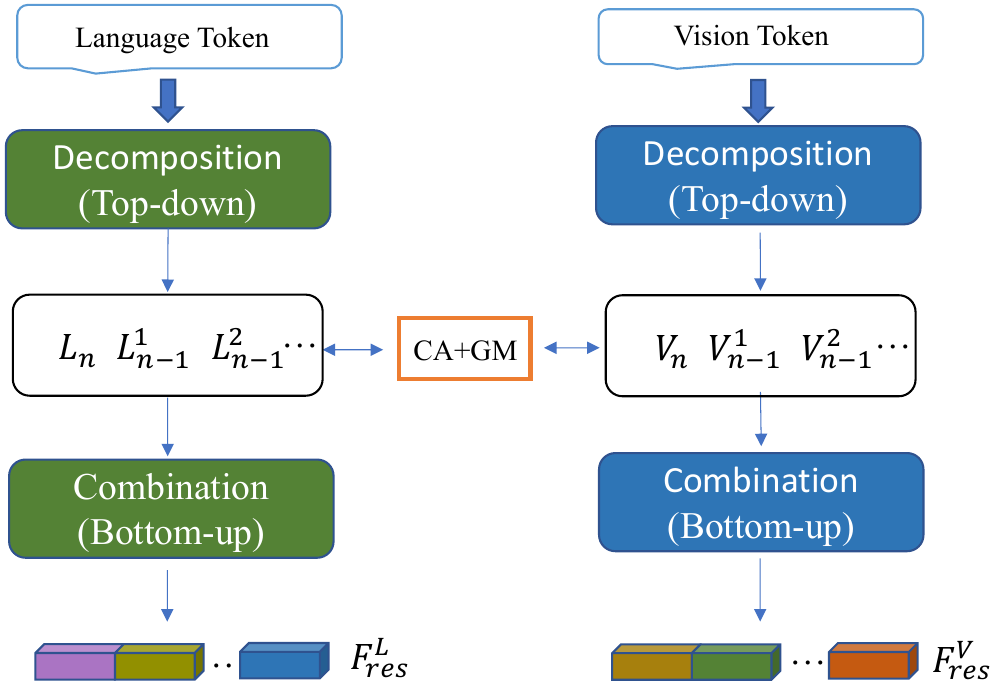}
	\caption{The specific design of PGVL: CA denotes Cross Attention and GM denotes Guided Module. Visual and language tokens are decomposed separately to obtain node features in their respective parse graphs. Starting from the leaf nodes, Cross Attention (CA) filtered by the Guided Module (GM) is performed between the corresponding node features of the two parse graphs. Then, it recursively proceeds upwards to the root node to obtain new features $F^L_{res}, F^V_{res}$.}
	\label{fig3}
\end{figure}

{\bf{In the top-down decomposition.}} As shown in Fig.~\ref{fig1}, the language token $F^L$ and visual token $F^V$ are decomposed along the channel separately to obtain node features in the parse graph, starting from the root node and proceeding towards the leaf nodes. The decomposition process is controlled by a parameter $\mathcal{G}$, where $\mathcal{G}=[g_n,g_{n-1},...,g_1]$ and $n$ represents the level of the tree structure (with leaf nodes at level 0 and the root node at level $n$).  Where $n\geq1$, each element $g_k   (k=1,2,...,n)$
indicates the number of sub-feature maps generated by decomposing a node at level $k$ into its children at level 
$k-1$. Let the channel number of the root node (at level $n$) be $C$. When a node at level $k$ with channel number $C_k$ is decomposed into $g_k$ sub-features at level $k-1$, the channel number of each sub-feature map at level $k-1$ is $C_{k-1}=\frac{C_k}{g_k}$. Finally, the visual and language node features are represented respectively (see Fig.~\ref{fig3}):
\begin{align}
	\label{eq1}
	N_L&=\{L_n, L^1_{n-1}, L^2_{n-1},  \cdots\} 	\\
	\label{eq2}
	N_V&=\{V_n, V^1_{n-1}, V^2_{n-1},  \cdots\} 
\end{align}
where the $N_L$ and $N_V$ represent the feature sets of all node in the corresponding parse graph. 
In the set, $n$ represents the level of the parse graph, from root node to leaf nodes, also from $n$ to $0$. The subscript of each element in the set indicates the level, and the superscript indicates the order of nodes at the same level (the root node has no superscript). 

\begin{figure}[t]
	\centering
	\includegraphics[width=0.98\linewidth]{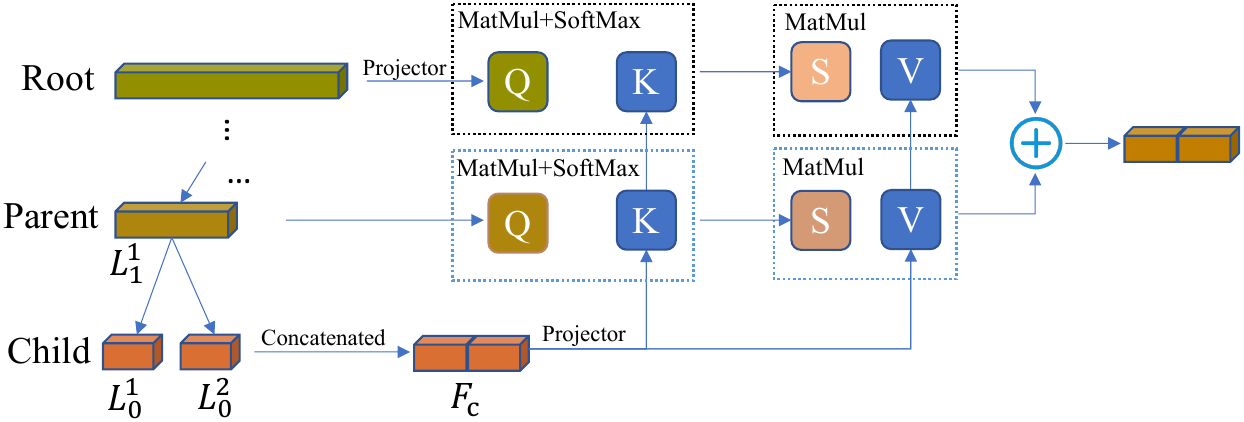}
	\caption{The specific design of Guided Module. The child nodes with cross-attention and context information are concatenated and then filtered using the parent node and the root node.}
	\label{fig4}
\end{figure}

{\bf{In the bottom-up combination.}} The context relations among nodes within each parse graph represented by $\mathcal{C}_{L^o_{k}}$ and $\mathcal{C}_{V^o_{k}}$.
 Let $\mathcal{T}_{L^o_k}$ or $\mathcal{T}_{V^o_k}$ (e.g., $\mathcal{T}_{L^1_0}=\mathcal{T}_{L^2_0}=\{L^1_0,L^2_0\}$ in Fig.~\ref{fig4}) denote the set of all nodes sharing the same parent. Assume that the size of each element in $\mathcal{T}$ is $(S,Ch)$, where $S$ is the flattened spatial dimension and $Ch$ is the number of channels. Concatenating the child nodes along the along $S$ dimension yields $Concat(\mathcal{T})\in(S*num,Ch)$, where $num$ is the number of child nodes. The context relations of each child node are then calculated:
\begin{align}
 		\label{eq-1}
 		\{\mathcal{C}_{L^o_{k}}\}_{L^o_k\in \mathcal{T}_{L^o_k}}&=Attention(Concat(\mathcal{T}_{L^o_k}),Concat(\mathcal{T}_{L^o_k}),Concat(\mathcal{T}_{L^o_k}))\\
 		\label{eq-2}
 		\{\mathcal{C}_{V^o_{k}}\}_{L^o_k\in \mathcal{T}_{V^o_k}}&=Attention(Concat(\mathcal{T}_{V^o_k}),Concat(\mathcal{T}_{V^o_k}),Concat(\mathcal{T}_{V^o_k}))
\end{align}
where $\mathcal{C}_{L^o_k}, \mathcal{C}_{V^o_k}\in R^{S\times Ch}$ are correspond to $\mathcal{N}$ in Eq.~\ref{eq0} and $Attention$ is the attention mechanism of Transformer~\cite{AttentionIsAllYouNeed}, where the first, second, and third terms correspond to $Q$, $K$, and $V$ respectively:
\begin{equation}
	\begin{split}
		\label{eq5}
		Attention(Q,K,V)=Softmax(\frac{QK^T}{\sqrt{d_k}})V
	\end{split}
\end{equation}
where $Q$ is the query matrix, $K$ is the key matrix, $V$ is the value matrix, $d_k$ is the dimension of the feature map.

Then cross attention is performed between corresponding nodes of the two parse graphs from leaf to root to complete visual-language fusion:
\begin{align}
	\label{eq3}
	M_{L^o_{k}}&=Attention (L^o_k,V^o_k,V^o_k)	\\
	\label{eq4}
	M_{V^o_{k}}&=Attention (V^o_k,L^o_k,L^o_k)
\end{align}
where $M_{L^o_{k}}$ is the cross modality information of node ${L^o_{k}}$ in the parse graph of language token and $M_{V^o_{k}}$ is the cross attention information of node ${V^o_{k}}$ in the parse graph of vision token. 

Subsequently, the context (e.g., $\mathcal{C}_{L^o_{k}}$) and cross modality (e.g., $M_{L^o_k}$) information of node are added to obtain $X_{L^o_k}$. For a parent node $L^o_m$ at level $m (0<m\leq n)$, the elements in the set of child node with cross and context information $\mathcal{X}_{L^o_m}=\{X_{L^i_{m-1}}|i=1,2,\dots\}$ (e.g., $\mathcal{X}_{L^1_1}=\{X_{L^i_0}|i=1,2\}$ in Fig.~\ref{fig4}) are concatenated and then filtered by the GM to generate new features, which replace the original features of the parent node $L^o_m$. This recursive process follows a bottom-up logic, where each parent node's feature is updated based on its child nodes' processed features. This process is described by the following equations:
\begin{align}
	\label{eq6}
	{{L^o_{m}}}&=GM(\mathcal{X}_{L^o_m},A_{L^o_{m}})	\\
	\label{eq6_1}
	{{V^o_{m}}}&=GM(\mathcal{X}_{V^o_m},A_{V^o_{m}})	
\end{align}
where the result of GM is used to replace ${L^o_{m}}$ or ${V^o_{m}}$ in the top-down decomposition and $A_{L^o_{m}}$ or $A_{V^o_{m}}$ is a set consisting of the corresponding root node $V_n$ or $L_n$ and parent node $L^o_{m}$ or $V^o_{m}$ in the top-down decomposition. 

GM as shown in Fig.~\ref{fig4}, in the bottom-up recursive process, taking the parse graph of language token as an example, the child nodes at level 0 have completed the computation of context (e.g., $\mathcal{C}_{L_0^{1}}$,$\mathcal{C}_{L_0^{2}}$) and cross-modal (e.g., ${M}_{L_0^{1}}$, ${M}_{L_0^{2}}$) information, yielding $X_{L_0^1},X_{L_0^2}$. $L_0^1,L_0^2$ nodes belong to the same parent node $L^1_1$, and the parent node $L^1_1$ has only these two child nodes. After $X_{L_0^1}$ and $X_{L_0^2}$ being concatenated along the channel dimension, the result $F_c$ match the parent node in size, while higher-level nodes remain unprocessed. The parent node and root node use $Attention()$ in Eq.~\ref{eq5} to query the most similar features respectively from $F_c$:
\begin{align}
	\label{eq6_2}
	Q_{L_r}&=Attention(\mathcal{L}(L_n),F_c,F_c)	\\
	\label{eq6_3}
	Q_{L_p}&=Attention(L^1_1,F_c,F_c)	
\end{align}
where $Q_{L_r}$ the result of the root node query, $Q_{L_p}$ the result of the parent node query, and $\mathcal{L}$ is a linear layer used to adjust the dimension to be consistent with $F_c$ or $L_1^1$. Then the output of GM can be obtained:
\begin{align}
	\label{eq6_4}
	GM(\mathcal{X}_{L^1_1},A_{L^1_1})=Q_{L_r}+Q_{L_p}
\end{align}
where $\mathcal{X}_{L^1_1}=\{X_{L^1_0},X_{L^2_0}\},A_{L^1_1}=\{L^1_1,L_n\}$. This process is carried out recursively, starting from the leaf nodes (level 0) and proceeding layer by layer upwards until reaching the root node. The bottom-up combination process of visual parse graph is the same as language parse graph. Finally, the optimized node features are recursively combined into new language token $F^L_{res}$ and vision token $F^V_{res}$ in Fig.~\ref{fig3}.

{\bf{PGVL variants.}} Visual tokens and language tokens are mapped for $t$ times through $MLP$:
\begin{align}
	\label{eq7}
	\{L_{d_1},L_{d_2},...,L_{d_t}\}&=MLP(L)	\\
	\label{eq8}
	\{V_{d_1},V_{d_2},...,V_{d_t}\}&=MLP(V)	
\end{align}
where $V$ and $L$ are the encoded visual and language tokens respectively. Let $D=\{d_1,...,d_t\}$ represent the set of corresponding channel numbers for different semantic spaces. Then $\{L_{d_1},L_{d_2},...,L_{d_t}\}$ and $\{V_{d_1},V_{d_2},...,V_{d_t}\}$ denote the sets of $L$ and $V$ mapped to multiple semantic spaces via MLP, respectively, where each element represents the feature. Subsequently, $L_{d_i}$ and $V_{d_i}$ in the two sets are processed by PGVL:
\begin{align}
	\label{eq9}
	F_{res}^{L_{d_i}},F_{res}^{V_{d_i}}=PGVL(L_{d_i},V_{d_i}),\quad         	i\in0,1,...,t
\end{align}
where $F_{res}^{L_{d_i}}$ and $F_{res}^{V_{d_i}}$ are the results after PGVL processing, that is, $F_{res}^L$ and $F_{res}^V$ in Fig.~\ref{fig3}. Next, visual and language results in multiple semantic spaces are concatenated respectively:
\begin{align}
	\label{eq10}
	F_{res}^{L_{Multi.}}&=Cat(F_{res}^{L_{d_0}},F_{res}^{L_{d_1}},...,F_{res}^{L_{d_t}}) \\
	\label{eq11}
	F_{res}^{V_{Multi.}}&=Cat(F_{res}^{V_{d_0}},F_{res}^{V_{d_1}},...,F_{res}^{V_{d_t}}) 
\end{align}
where $Cat$ is a function that we use to concatenate the fusion results of different semantic spaces along the channel.

Finally, residuals are added to the output of PGVL or its variants to ensure that the channel performance of the original feature maps is preserved.

\subsection{VLML} 
Let the output of PGVL or its variants is $F_{res}^V,F_{res}^L$. $F_{res}^V$ using the ground truth position of joints to obtain the local visual features of $K$ joints, $F_n\in N^{1\times C}, n\in0,1,...,K$. Then the stacked joints feature $F_{joints}$ can be obtained. Next, the correlation matrix of visual and language features can be obtained:
\begin{align}
	\label{eq12}
	Corr=F_{joints}\times T(F_{res}^{L}),\quad Corr\in R^{K\times K}
\end{align}
where the shape of $F_{res}^{L}$ and $F_{joints}$ are $N^{K\times C}$, $T$ is the matrix transpose, and $\times$ is matrix multiplication.

Given that each visual keypoint feature has a unique prompt embedding, we can use a diagonal matrix as the matching target $Corr_{label}$. Finally, We apply contrastive learning to both prompt embeddings and keypoint features using the following feature-level contrastive loss:
\begin{align}
	\label{eq13}
	VLML=0.5[&CEL(Corr,Corr_{label})+ \\
	\notag
	&CEL(Corr^T,Corr_{label})]
\end{align}
where $T$ is the matrix transpose and $CEL$ represents the cross entropy loss.

\section{Experiments}

\begin{table}[!t]	
	\centering
	\begin{tabular}{c|ccccccc|c}
		\hline
		Method  & Head & Sho. & Elb. & Wri. & Hip  & Knee & Ank. & Mean \\ \hline
		\multicolumn{9}{c}{Multi-scale testing} \\ \hline
		Newell et al.~\cite{Hourglass}& 97.1 & 96.1& 90.8 & 86.2 & 89.9& 85.9& 83.5 &  90.0\\
		Yang et al.~\cite{FPN}& 97.4& 96.2& 91.1 & 86.9 & 90.1& 86.0& 83.9&  90.3\\
		Tang et al.~\cite{DeeplyLearnedCompositionalModels}& 97.4& 96.2& 91.0& 86.9 & 90.6& 86.8& 84.5&  90.5\\
		Sim.Bas.~\cite{SimpleBaseline}& 97.5 & 96.1  & 90.5 & 85.4 & 90.1& 85.7& 82.3&  90.1\\
		
		HRNet-W32~\cite{Hrnet}& 97.7& 96.3  & 90.9& 86.7& 89.7& 87.4& 84.1&  90.8\\
		PGBS~\cite{PGBS}& 97.4& 96.5  & 91.8& 87.6& 90.3& 87.9& 84.1&  91.2\\ 
		RMPG(large)~\cite{RMPG}& 97.2& 96.4  & 91.4& 87.5& 90.3& 87.9& 84.4&  91.2\\ \hline
		\multicolumn{9}{c}{Single-scale testing} \\ \hline
		
		Tang et al.~\cite{DeeplyLearnedCompositionalModels}& 95.6& 95.9& 90.7& 86.5& 89.9& 86.6& 82.5 &  89.8\\
		Sim.Bas.~\cite{SimpleBaseline}& 97.0 & 95.9  & 90.3 & 85.0 & 89.2& 85.3& 81.3  &  89.6\\
		ViTPose-B*~\cite{Vitpose}& 96.7 & 96.0 & 90.2& 85.5 & 89.0& 86.8& 82.3 &  90.0\\
		HRNet-W32& 97.1 & 95.9     & 90.3  & 86.4  & 89.1 & 87.1 & 83.3  &  90.3\\  \hline
		Ours (ViT-B)       & 97.4& 96.7& 91.9& 87.9& 91.3& 88.6& 84.9  & {\bf{91.7}}\\ 
		\hline
	\end{tabular}
	\caption{Comparisons on the MPII val set (PCKh@0.5). * denotes our replicated results.}
	\label{tb1}
\end{table}

\begin{table}[t]	

	\centering
	\begin{tabular}{c|c|c|c}
		\hline
		Method     & Backbone &Input size& MAP   \\ \hline
		SimpleBaseline~\cite{SimpleBaseline}& ResNet-101 &$320\times256$    & 66.1 \\
		MIPNet~\cite{Mipnet}& ResNet-101 &$384\times288$    & 68.1 \\
		RTMO~\cite{RTMO}& RTMO-s &$640\times640$    & 67.3\\
		DEKR~\cite{DEKR}& HRNet-W32 &$512\times512$    & 66.3\\
		SimpleBaseline~\cite{SimpleBaseline}& ResNet-50 &$256\times192$    & 63.7 \\
		SimpleBaseline~\cite{SimpleBaseline}& ResNet-101 &$256\times192$    & 64.7 \\
		SimpleBaseline~\cite{SimpleBaseline}& ResNet-152 &$256\times192$    & 65.6 \\
		HRNet~\cite{Hrnet}& HRNet-W32 &$256\times192$    & 67.5  \\
		VITPose*~\cite{Vitpose}& ViT-B&$256\times192$   & 66.3   \\ \hline
		Ours& ViT-B&$256\times192$   & {\bf{68.2}}\\ 
		
		\hline
	\end{tabular}
	\caption{Comparisons on the CrowdPose test set with a YOLOv3 human detector~\cite{Yolov3}. * means our reproduce results.\label{tb2}}
\end{table}

\subsection{Datasets and Evaluation Methods}

{\bf{Datasets.}} The performance of PGVL and our network is evaluated using the MPII~\cite{Mpii}, CrowdPose~\cite{Crowdpose},  OCHuman~\cite{OCHuman},AP-10K~\cite{AP10K} and Animal-Pose~\cite{AnimalPose} datasets. The CrowdPose dataset, composed of 20,000 images and 80,000 human instances each annotated with 14 keypoints, is divided into training, validation, and testing subsets at a 5:1:4 ratio. The MPII Human Pose dataset, containing approximately 25,000 images and 40,000 annotations with 16 keypoints each, includes 28,000 training and 11,000 testing samples. The OCHuman dataset, which includes human instances with severe occlusion and 4K images and 8K instances, is exclusively used for validation and testing. For the animal pose datasets, the AP-10K dataset, the largest and most diverse for animal pose estimation, includes 10,015 images from 23 animal families and 54 species, with 17 keypoints annotated. The Animal-Pose dataset covers 5 different animal species with over 4,000 images. 20 keypoints are annotated in each animal instance.

For the OCHuman dataset, we train our network using the COCO keypoint detection dataset~\cite{COCO} with results reported on the OCHuman validation set. For other datasets, we train the network with their corresponding training sets and report the results on their validation or test sets.

{\bf{Evaluation methods.}} For the MPII dataset, the PCKh score is used as evaluation indicators. For other datasets, we use Mean Average Precision (MAP) and Mean Average Recall (MAR) as evaluation indicators
\subsection{Implementation Details}

{\bf{Training and testing.}} In our experiments, our network adopts a top-down pose estimation approach similar to HRNet~\cite{Hrnet}. For pre-trained model selection, the visual backbone uses an ImageNet~\cite{Imagenet}-pre-trained model, while the language backbone employs a CLIP~\cite{CLIP}-pre-trained model. Each model is trained for 210 epochs with a stepwise learning rate scheduler that decays the learning rate by a factor of 10 at the 170th and 200th epochs, and the AdamW optimizer~\cite{Adamw} is applied.

{\bf{The settings of $\mathcal{G}$ and $D$.}} Unless otherwise specified, for all experiments, including ablation studies and visualization results, the decomposition process control parameter $\mathcal{G}=\{2,2,2\}$ and the channel number for different semantic spaces $D=\{512\}$, which is a single semantic space.

\begin{table}[t]	

	\centering
	\begin{tabular}{c|c|c|c}
		\hline
		Method     & Backbone & Input size&MAP   \\ \hline
		
		SimpleBaseline~\cite{SimpleBaseline}& ResNet-152 &$384\times288$    & 58.8 \\
		HRNet~\cite{Hrnet}& HRNet-W32 &$384\times288$     & 60.9 \\
		HRNet~\cite{Hrnet}& HRNet-W48 &$384\times288$   & 62.1\\
		HRFormer~\cite{HRFormer}&HRFormer-S &$384\times288$    & 53.1  \\
		HRFormer~\cite{HRFormer}&HRFormer-B &$384\times288$    & 50.4  \\ 
		VITPose*~\cite{Vitpose}& ViT-B&$256\times192$   & 60.4 \\ \hline
		Ours& ViT-B&$256\times192$    & {\bf{62.8}}\\ 
		\hline
	\end{tabular}
	\caption{Comparisons on the OCHuman val set. * means our reproduce results.\label{tb3}}
\end{table}

\subsection{Benchmark Results and PGVL Efficacy in Other Methods}
{\bf{MPII benchmark.}} Table~\ref{tb1} shows the results of our network and existing advanced methods on the MPII val set, where the input of all methods is $256\times256$. Our network with the backbone ViT-B achieves 91.7 Mean (PCKh@0.5), 0.5 highter than the methods of RMPG~\cite{RMPG} and PGBS~\cite{PGBS} and 1.7 higher than the ViTPose-B~\cite{Vitpose}.

{\bf{CrowdPose benchmark.}} Table~\ref{tb2} shows the results of our network and existing advanced methods on the CrowdPose test set. Our network with the backbone ViT-B achieves 68.2 MAP, 0.7 higher than the HRNet-W32~\cite{Hrnet} and 1.9 higher than the ViTPose~\cite{Vitpose} with the backbone ViT-B.

{\bf{OCHuman benchmark.}} Table~\ref{tb3} shows the results of our network and existing advanced methods on the OCHuman val set. Our network with the backbone ViT-B achieves 62.8 MAP, 0.7 higher than HRNet-W48 and 2.4 higher than the ViTPose with the backbone ViT-B.

\begin{table}[!t]	

	\centering
	\begin{tabular}{c|c|c|c|c}
		\hline
		Method     & Backbone &Pre-train & MAP  & MAR \\ \hline
		ViTPose*~\cite{Vitpose} & ViT-B&ImageNet & 73.5 & 76.6\\ 
		ViTPose*~\cite{Vitpose} & ViT-L&ImageNet & 76.0& 79.3 \\ 
		HRNet~\cite{Hrnet} & W32&ImageNet   & 73.8   & - \\ 
		HRNet~\cite{Hrnet} & W48&ImageNet   & 74.4   & - \\ 
		Sim.Base~\cite{SimpleBaseline} & ViT-B&CLIP  & 72.6   & 75.8 \\   
		CLAMP~\cite{CLAMP} & ViT-B&CLIP   & 74.3   & 77.5 \\  \hline
		CLAMP~\cite{CLAMP} & ViT-L&CLIP      & 77.8   & - \\ 
		Ours& ViT-L&ImageNet,CLIP  & {\bf{82.1}}  & 84.3 \\ 
		\hline
	\end{tabular}
	\caption{Comparisons on the AP-10K valid dataset. * is our reproduce results.\label{tb-1}}
\end{table}
{\bf{AP-10K benchmark.}} Table~\ref{tb-1} shows the results of our network and existing advanced methods on the AP-10K validation set, where the input of all methods is $256\times256$. Our network achieve $82.1$ MAP with the ViT-L backbone ($\mathcal{G}=\{2,2,2\}, D=\{512,768\}$), $4.3$ higher than the method of CLAMP~\cite{CLAMP} with the same backbone. 

{\bf{Animal-Pose benchmark.}} Table~\ref{tb-2} shows the results of our network and existing advanced methods on the Animal-Pose val set, where the input of all methods is $256\times256$. Our network achieve $79.3$ MAP with the ViT-B backbone ($\mathcal{G}=\{2,2,2\}, D=\{512,256\}$), $5.0$ higher than the method of CLAMP with the same backbone. 

\begin{table}[h]	

	\centering
	\begin{tabular}{c|c|c|c|c}	
		\hline
		Method     & Backbone &Pre-train& MAP  & MAR \\ \hline
		ViTPose*~\cite{Vitpose} & ViT-B&ImageNet & 70.2  & 74.1 \\ 
		Sim.Base.~\cite{SimpleBaseline} & ResNet-50&ImageNet & 69.1  & 73.6 \\ 
		Sim.Base.~\cite{SimpleBaseline} & ResNet-152&ImageNet & 70.4  & 74.8 \\ 
		HRNet~\cite{Hrnet} & W32&ImageNet & 74.0  & 78.0 \\ 
		HRNet~\cite{Hrnet} & W48&ImageNet & 73.8  & 77.8\\ 
		Sim.Base.~\cite{SimpleBaseline} & ResNet-50&CLIP     & 70.8 & 75.0 \\
		
		Sim.Base.~\cite{SimpleBaseline} & ViT-B&CLIP      & 72.3 & 76.3\\  \hline
		CLAMP~\cite{CLAMP} & ViT-B&CLIP  & 74.3   & 78.3 \\  
		Ours & ViT-B&ImageNet,CLIP     & {\bf{79.3}}& 82.7 \\ 
		\hline
	\end{tabular}
	\caption{Comparisons on the Animal-Pose valid dataset. * is our reproduce results\label{tb-2}}
\end{table}

{\bf{Network complexity analysis.}} As shown in Table~\ref{tb-3}, when our network employs PGVL with $\mathcal{G}=\{2,2,2\}$ and $D=\{512,256\}$, the complexity of our network is $192.2M$, $31M$ higher than that of CLAMP, but the MAP is 5.0 higher than CLAMP.
\begin{table}[!t]	

	\centering
	\begin{tabular}{c|c|c|cc}
		\hline
		Method     & Backbone & Parmas & MAP   \\ \hline
		CLAMP~\cite{CLAMP} & ViT-B& 161.2M    & 74.3    \\ 
		Ours & ViT-B & 192.2M  & 79.3    \\ 	\hline
	\end{tabular}
	\caption{Network parameter comparison on the Animal-Pose val set.\label{tb-3}}
\end{table}

{\bf{PGVL efficacy in other methods.}} Table~\ref{tb4} shows the results of introducing PGVL into other methods, with parameter increments from PGVL specified. {\bf{For LAMP~\cite{LAMP}}} (bottom-up multi-person pose estimation), language descriptions of joints and instances are introduced to enhance occlusion handling. Two PGVLs are used to separately fuse visual features with joint-language and instance-language description features, with $\mathcal{G}=\{2,2\}$ and $D=\{256\}$ for each. After introducing PGVL, AP increases by 0.3 and $AP^{50}$ increases by 0.6 and the number of parameters increase by $59.1M$ on OCHuman test set. {\bf{For CLAMP~\cite{CLAMP}}} (top-down single-animal pose estimation),  which combines visual and joint description features for pose estimation, a PGVL with $\mathcal{G}=\{2,2,2\}$ and $D=\{512\}$ are used to fused these features. After introducing PGVL, on the AP-10K val set, AP and $AP^{50}$ each increase by 1.0, with a $23.4M$ parameter increment. On the Animal-Pose val set, AP increases by 0.8, $AP^{75}$ by 0.5 and the parameter count also goes up by $23.4M$. 

Our network outperforms on diverse pose datasets, including those for animals, by leveraging text information to bridge dataset gaps and designing PGVL to fuse multimodal information. Compared with CLAMP, a related approach, our method demonstrates significant advantages.

\begin{table}[!t]	
	
	\centering
	\begin{tabular}{c|c|c|c|c|c}
		\hline
		
		Method& \makecell[c]{\#Params \\ on PGVL } & MAP& AP$^{50}$ & AP$^{75}$&MAR\\ \hline
		
		\multicolumn{5}{c}{OCHUMAN test set} \\ \hline
		LAMP*~\cite{LAMP} &-&58.2&  76.1&63.8&75.7\\ 
		LAMP+PGVL&59.1M&58.5&76.7&64.0&76.1\\ \hline
		\multicolumn{5}{c}{AP-10K val set} \\ \hline 
		CLAMP~\cite{CLAMP} &-&74.3&  95.8&81.4&77.5\\ 
		CLAMP+PGVL &23.4M&75.3&  95.8&82.9&78.5\\  
		\hline
		\multicolumn{5}{c}{Animal-Pose val set} \\\hline 
		CLAMP~\cite{CLAMP}&-&74.3&  95.8&83.4&78.3\\ 
		CLAMP+PGVL &23.4M&75.1&  95.7&83.9&79.0\\
		\hline
		
	\end{tabular}
	\caption{The experiment of PGVL in other method. * is our reproduce results. \label{tb4}}
\end{table}

\begin{figure}[t]
	\centering
	\includegraphics[width=1\linewidth]{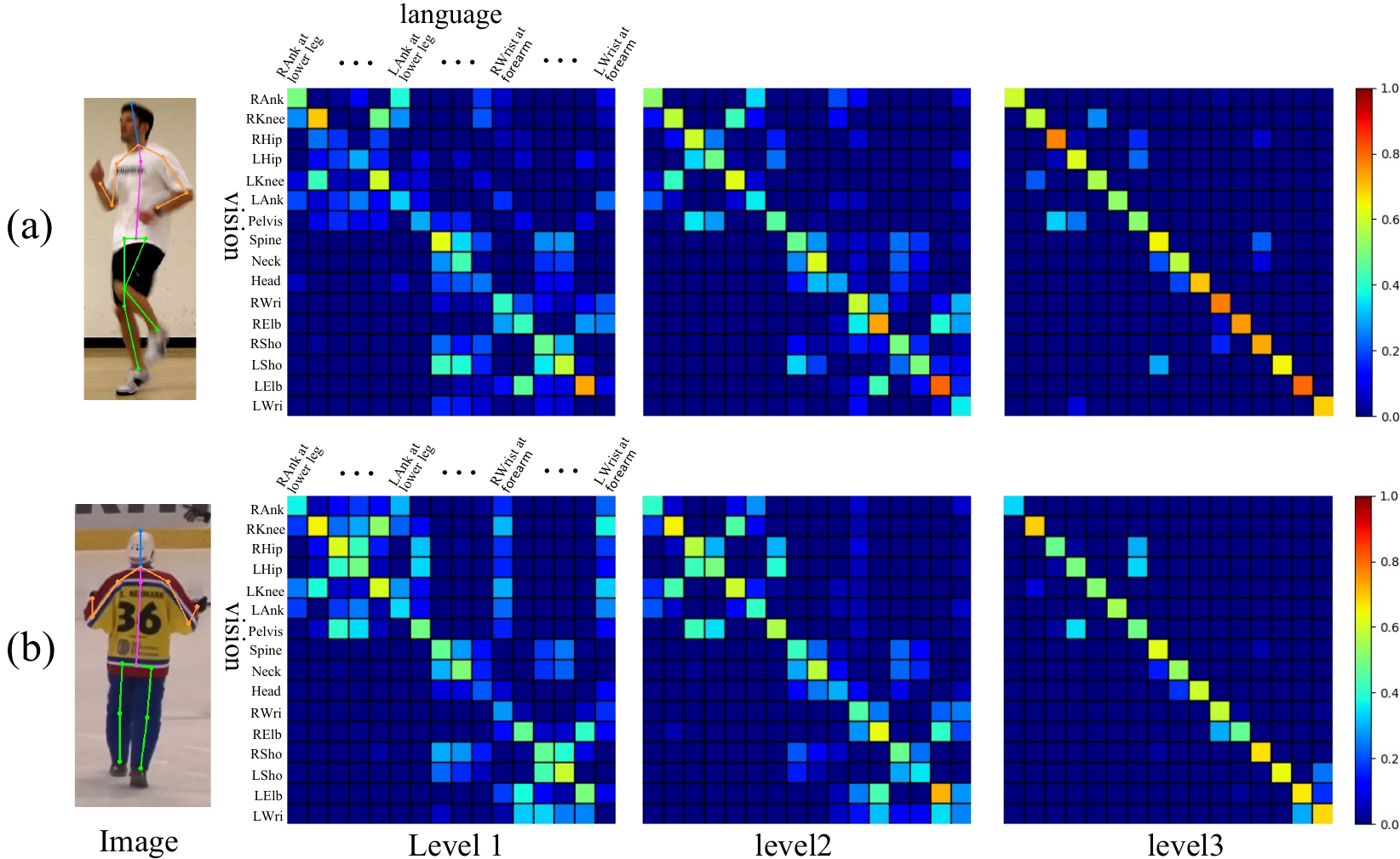}
	\caption{Examples (a) and (b) demonstrate the hierarchical visual-language alignment results of the PGVL module in our network, which are, in sequence: the input image, the alignment features of the first layer (low-level), the second layer, and the third layer (high-level).}
	\label{fig5}
\end{figure}

\begin{figure}[!t]
	\centering
	\includegraphics[width=1\linewidth]{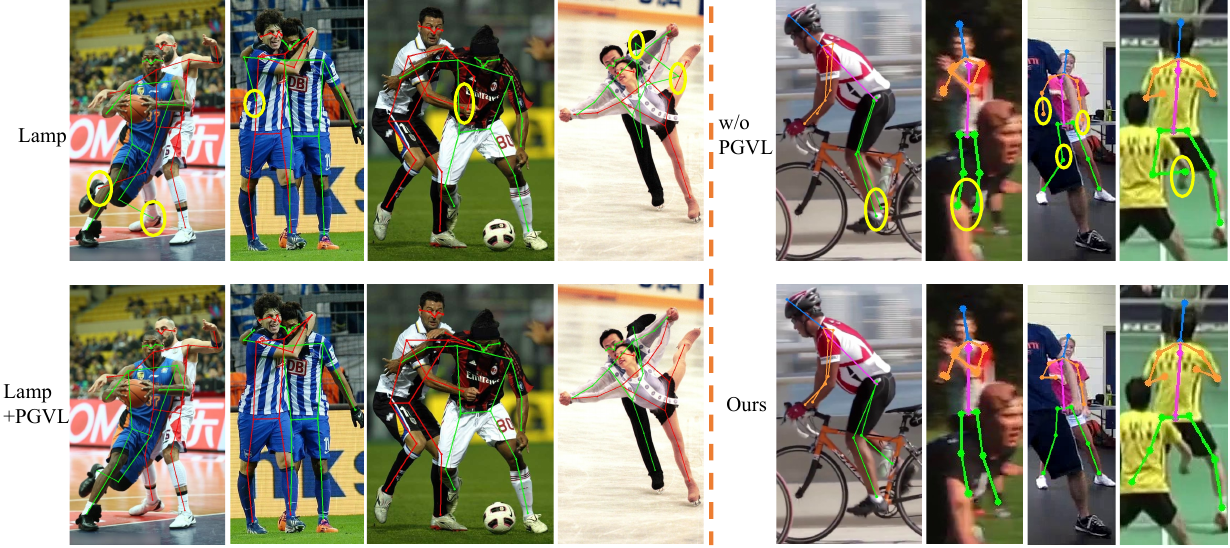}
	\caption{The vertical orange dashed line demarcates: (left) multi-person pose estimation comparisons between LAMP and LAMP+PGVL on OCHuman test set; (right) single-person results comparing the ablated version (without PGVL) against our full network on MPII val set.}
	\label{fig6}
\end{figure}

\begin{figure}[t]
	\centering
	\includegraphics[width=1\linewidth]{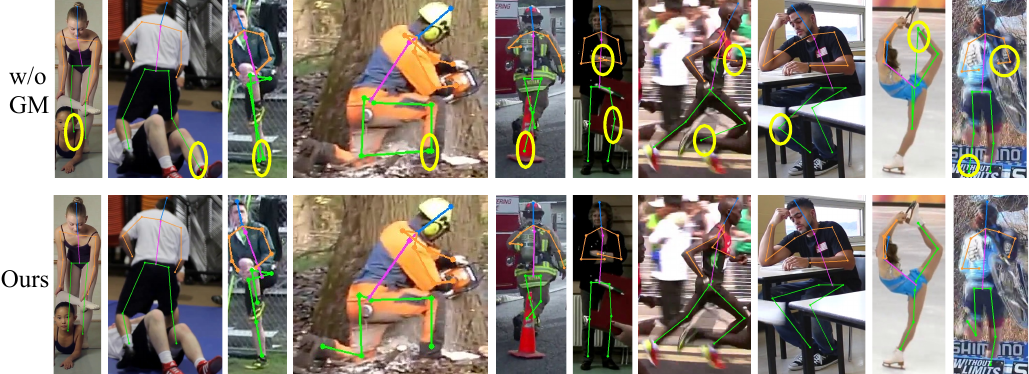}
	\caption{Single-person results comparing the ablated version (without GM) against our full network on MPII val set.}
	\label{fig7}
\end{figure}

{\bf{Visualization of PGVL.}} To further demonstrate the visual-language fusion effect of PGVL, we visualize the visual-language alignment results at each level of PGVL with $\mathcal{G}=\{2,2,2\}$ and $D=\{512\}$ in our network, as shown in Fig.~\ref{fig5}. In the first example, when the right knee of a person is occluded by the left knee, the right knee can still achieve high-similarity alignment with its description at level 1 and maintain this effect in subsequent levels, reflecting the advantage of local information. At level 2 and level 3, global optimization gradually reduces the scores of irrelevant descriptions, demonstrating the role of global constraints. In the second example, the left wrist is invisible, leading to alignment failure and lack of relevant features at level 1. However, at level 2 and level 3, by leveraging increasing global information, the position of the left wrist is gradually inferred and aligned with its description, highlighting the value of global information.

\begin{table}[!t]	

	\centering
	\begin{tabular}{c|c}
		\hline
		Method & \makecell[c]{MPII val \\Mean } \\ \hline
		Base (w/o PGVL)   & 90.8\\
		PGVL w/o Context relations       & 91.6\\
		
		PGVL w/o Cross-Attention       & 91.5\\  
		PGVL w/o Guided Module     & 91.3\\  
		\hline
		PGVL (Prompt w/o direction)  & 91.5 \\
		PGVL & 91.7 \\ 
		\hline
	\end{tabular}
	\caption{Ablation experiment. Prompt without direction means changing "left ankle" to "ankle", and the prompts for other joints follow the same logic. \label{tb5}}
\end{table}

\begin{table*}[!t]
	
	\centering
	\begin{tabular}{c|c|c|c|c|}
		\hline
		\multirow{2}{*}{Method} & \multicolumn{2}{c|}{\textbf{PGVL Variants}} &\multirow{2}{*}{\makecell[c]{\#Params \\ \textbf{Fuse only}}}& \multirow{2}{*}{\makecell[c]{ MPII val \\ Mean}} \\
		\cline{2-3}
		& $\mathcal{G}$ &$D$ &  & \\
		\hline
		\multirow{6}{*}{\makecell[c]{Our Network\\(ViT-B)}}
		& \multirow{2}{*}{$\{2,2\}$}
		& $\{512\}$          & 20.0M & 91.5 \\
		& & $\{512,256\}$ & 31.3M &91.7 \\
		\cline{2-5}
		& \multirow{3}{*}{$\{2,2,2\}$} 
		& $\{256\}$          & 5.7M & 91.5\\
		&& $\{512\}$          & 22.9M & 91.7\\
		& & $\{512,256\}$ & 34.9M & {\bf{91.8}}\\
		
		\cline{2-5}

		\cline{2-5}
		&\multicolumn{2}{c|}{\textbf{Cross-Attention~\cite{AttentionIsAllYouNeed}}} &63.0M&91.4\\ \hline
	\end{tabular}
	\caption{PGVL variants experiments. Cross-Attention~\cite{AttentionIsAllYouNeed} is a fundamental fusion method and used in the LAMP~\cite{LAMP} method \label{tb6}}
\end{table*}

{\bf{Visualization of final results.}} We visualize the final results, as shown in Fig.~\ref{fig6}. The left side compares the visualizations of LAMP and LAMP with PGVL in bottom-up multi-person pose estimation, while the right side compares the visualizations of our network with and without PGVL in top-down single-person pose estimation. In addition, as shown in Fig.~\ref{fig7}, we visualized the final results of the network, including the PGVL with GM and the PGVL without GM in the network. The results demonstrate the effectiveness GM.

\subsection{Ablation Study}
{\bf{PGVL analysis.}} Our network employs PGVL with $\mathcal{G}=\{2,2,2\}$ and $C=\{512\}$. It's ablation experiments are made on the MPII val set with the input size of $256\times256$. Table~\ref{tb5} shows the results of our ablative experiment. Ablation shows removing PGVL's context relation, cross-attention, and GM decreases PCKh by 0.1, 0.2, and 0.4 respectively, indicating each component is effective—particularly the GM. 

{\bf{Prompt Effectiveness.}} As shown in Table~\ref{tb5}, we remove directional words from the language descriptions of joints (e.g., changing 'left ankle' to 'ankle') and found that the index decreased by 0.2 PCKh, demonstrating the importance of orientation in language description, which can provide more directional guidance for vision.

{\bf{PGVL variants.}} Table~\ref{tb6} presents additional experiments on PGVL variants. When $\mathcal{G}=\{2,2\}$, adjusting $D$ from $\{512\}$ to $\{512,256\}$ increases PCKh by 0.2. For $\mathcal{G}=\{2,2,2\}$, increasing $D$ from $\{256\}$ to $\{512\}$ boots the metric by 0.2, and further adjusting $D$ to $\{512,256\}$ enhances it by an additional 0.1. These results demonstrate the effectiveness of multi-scale semantic spaces and appropriate channel increases. Notably, all these methods outperform the baseline fusion approach in metrics while requiring fewer parameters.

\section{Conclusion}

We propose a novel visual-language fusion method based on parse graphs (PGVL) to address visual-language alignment and location under occlusion, and construct a new pose estimation network framework. The effectiveness of PGVL and our network framework has been validated. We hope that our method and the techniques in PGVL, including hierarchical multimodal fusion and Guidance Modules based on high level nodes, can be more widely used in the field of artificial intelligence.


%
%
%

\bibliographystyle{elsarticle-num} 
\bibliography{PGVL}



%
%
%
\end{document}